# Texture image analysis based on joint of multi directions GLCM and local ternary patterns

**Akshakhi Kumar Pritoonka[1], Faeze Kiani[2,*]**

[1]*Department of textural science, Online Computer Vision Research Group, India*
[2]*Department of electronic science, Online Computer Vision Research Group, Iran*
*\*ocvrgroup000@gmail.com*

**Abstract:** Human visual brain use three main component such as color, texture and shape to detect or identify environment and objects. Hence, texture analysis has been paid much attention by scientific researchers in last two decades. Texture features can be used in many different applications in commuter vision or machine learning problems. Since now, many different approaches have been proposed to classify textures. Most of them consider the classification accuracy as the main challenge that should be improved. In this article, a new approach is proposed based on combination of two efficient texture descriptor, co-occurrence matrix and local ternary patterns (LTP). First of all, basic local binary pattern and LTP are performed to extract local textural information. Next, a subset of statistical features is extracted from gray-level co-occurrence matrixes. Finally, concatenated features are used to train classifiers. The performance is evaluated on Brodatz benchmark dataset in terms of accuracy. Experimental results show that proposed approach provide higher classification rate in comparison with some state-of-the-art approaches.

## 1. Introduction

Human visual brain uses three main components such as color, texture and shape, to detect or identify environment and objects. Color, texture, shape, and geometric features are the main components that the human visual system uses to identify the environment and objects. In about 65% of cases, texture plays the most important role compared to color and shape in recognizing objects and content [1]. Therefore, texture classification has received a lot of attention from scientific researchers in the last decade. The texture function can be used in many different passenger vision and machine learning problems, such as medical image recognition [2-4], image retrieval [5-7], object recognition [8], skin recognition [9], biochemistry [10], defect detection [11,12], Bioinformatics [13], etc. In the last decade, texture classification and texture analysis are two of the main topics in the literature of computer vision and image processing science. Since texture classification is closely related to sciences like machine learning, it works in areas like pattern recognition, object tracking, defect detection, face tracking, etc. The main problem in texture classification is related to two issues:

  I.     The best characteristics for the description of textures

  II.    Choose the best and most suitable type of classifiers with the selected characteristics.

Regarding these two issues, we have witnessed several discussions. Some of these approaches work directly on images taken from textures, such as texture classification based on the random threshold vector technique [14] and texture classification based on primitive pattern units [15]. Another group of approaches first processes images and then looks for suitable features related to class labels, such as texture classification using advanced local binary patterns and spatial distribution of dominant patterns [16] and a new color texture for Industrial products inspection approach [17, 18]. Since the approaches presented so far are practical and suitable for each case, they are not guaranteed to work effectively in other applications. For this reason, the main objective of the researcher is to define new approachs and characteristics. This article presents a new approach that works in a general way. It can be used for various applications, but it works well on the rock texture classification problem and can accurately classify all kinds of rock textures using a train step. In this approach, the image is first processed using the LBP and GLCM algorithms. Then, using edge filters, edge detection has been performed on the image. We can build a suitable data set by extracting well-known statistical features, such as energy, entropy, and contrast. Finally, by using classifiers, we can classify the test data with high accuracy. The reminder of this article is as follows: The section is related to the literature review. The LBP, LTP and GLCM process is explained in this section. In the third section, the proposed texture classification approach is clearly explained in detail. The experimental results are included in the fourth section. Finally, conclusions and future work are presented in Section 5.

## 2. Related works

As mentioned above, Texture image analysis can be performed using different operations. In this article, local ternary patterns and gray-level co-occurrence matrixes are employed to extract texture features. So, the process structure of these operators is described with details in this section.

### 2.1. Local ternary patterns

LTP is an efficient texture analysis which can be used in different applications such as. LTP can be considered as one of the efficient versions of the local binary pattern (LBP). So, it is better to review LBP computation process firstly. LBP is one of the most popular texture analysis operators that were introduced first time in [19]. It is a gray-scale invariant texture measure computed from the analysis of an N×N local neighborhood over a central pixel. LBP generates a binary code by thresholding a local center neighborhood by the gray value of its center. Next, a label between {0, 1} is assigned to each one of the neighbors by comparing their values to the central pixel value. If the neighbor's intensity is below the intensity of the central pixel, then it is labeled 0, otherwise it is assigned the value 1:

$$P_i^{'} = \begin{cases} 0 & \text{if } I(x_i, y_i) < I(x_c, y_c) \\ 1 & \text{Otherwise} \end{cases} \qquad (1)$$

$P_i^{'}$ is the obtained binary label for $i^{th}$ neighbors, $I(x_i, y_i)$ is the intensity value of the $i^{th}$ neighbors. Also, $I(x_c, y_c)$ shows the intensity of the central pixel. The considered label of each neighbor is then multiplied by weights given to the corresponding pixels. The weight is given by the value $2^{i-1}$. Summing the obtained values gives the measure of the LBP. In Eq. 2, the parameter N is the total number of neighbors. An example on how to compute LBP is shown in the figure 1. The neighborhood size is 3×3 in the Fig. 1. An example of the weights kernel is shown in the Fig. 1.

$$LBP = \sum_{i=1}^{N} P_i^{'} 2^{i-1} \qquad (2)$$

| 8 | 22 | 12 |
|---|----|----|
| 93 | 50 | 55 |
| 8 | 89 | 5 |

| 0 | 0 | 0 |
|---|---|---|
| 1 | C | 1 |
| 0 | 1 | 0 |

| 32 | 64 | 128 |
|----|----|-----|
| 16 | C | 1 |
| 8 | 4 | 2 |

LBP(C) = 00010101   = 21

**Figure1.** Computation LBP example

In the figure 1, the sum of the resulting values gives the LBP measure is 21. The central pixel is replaced by the obtained value. By applying the LBP operator on the whole image, a new matrix is calculated. Basic version of the LBP is sensitive to impulse noise [20]. So, improved version LTP proposed for the first time in [21]. In LTP operator, for each pixel of an image a local ternary pattern is computed which is a 3-valued codes (-1, 0, 1). The LTPs is defined as follows:

$$LTP_{P,R} = \sum_{i=0}^{P-1} s(g_i - g_c) 3^i \quad , \qquad s(x) \begin{cases} 1 & \text{if} \quad x >= t \\ 0 & \text{if} \quad |x| < t \\ -1 & \text{if} \quad x <= -t \end{cases} \qquad (3)$$

Where, *t* denotes the user threshold for coding. Output of the $LTP_{P,R}$ operator for each pixel of the image is a P-bit binary number with $3^P$ different values. This increases the computation complexity so for simplicity and reduction of computational complexity, the ternary code is divided into two upper LBP and lower LBP. An example of LTP encoding procedure and split it into upper and lower patterns is illustrated in figure (2).

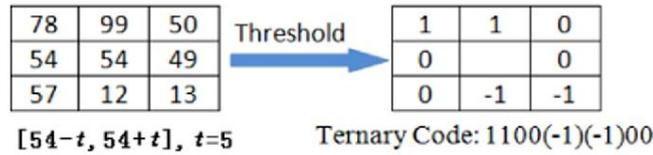

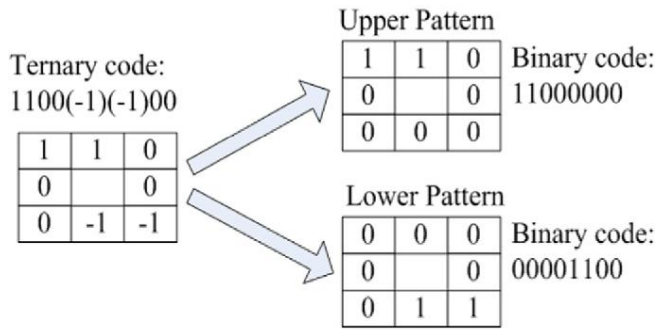

**Figure 2.** A visual example of LTP encoding procedure

In the end, the two upper and lower local binary patterns histograms are generated individually and connected together to produce a single histogram. LTP provide more discriminative features than LBP and some other version of LBP-based on reported results in recent articles [22-25].

## 2. 2. Gray Level Co-occurrence Matrix

The gray level coherence matrix (GLCM) was proposed in [26] by Haralik and Shanmugam. It is very useful in tissue analysis. Computes second-order statistics related to image features considering the spatial relationship of the pixels. GLCM shows how often different combinations of gray levels occur simultaneously in an image [27-29]. GLCM is created by computing every few times a pixel with intensity value i in a specific spatial relationship with a pixel with value j. The spatial relationship can be specified in several ways, the default being the relationship between a pixel and its immediate neighbor to the right [30]. However, we can specify this relationship with different offsets and angles. The pixel at position (i,j) in GLCM is the sum of the number of times the relationship (i,j) occurs in the image. Figure 3 describes how to calculate the GLCM. This displays an image and its corresponding co-occurrence matrix using the default pixel spatial relationship (offset = +1 in the x-direction). For the pair (2,1) (pixel 2 followed by pixel 1 to the right), 2 is found in the image, so the GLCM image will have 2 as the value at the position corresponding to Ii = 1 and Ij = . 2. The GLCM array is a 256x256 array. Ii and Ij are intensity values for an 8-bit image. GLCM can be calculated for eight directions around the desired pixel (Figure 3). The summation of the results from different directions leads to an isotropic GLCM and helps to achieve a rotationally invariant GLCM (Figure 4).

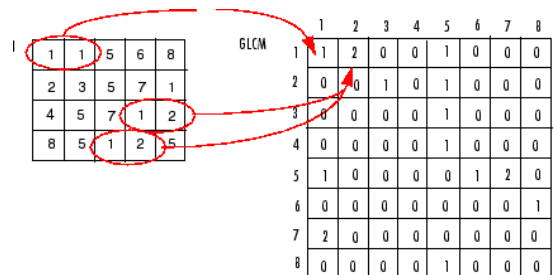

**Figure3**. Numerical GLCM procedure examples

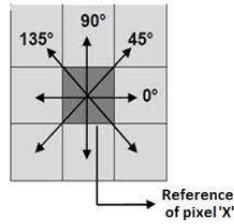

**Figure4.** Eight popular directions used to generate for computing isotropic GLCM.

## 3. Texture image classification using joint of LTP and GLCM

The proposed approach is a multistage approach. First of all, LBP operation is assigned to the whole image. Histogram of the output LBP-filtered image is considered as a feature vector. Next, LTP is performed which provide two different feature vectors with $2^N$ dimensions. Where, N shows the total number of neighbors based on radius of LBP neighborhood. For example, if a neighborhood with size 3×3 is considered, finally, LBP histogram has 256 bins. So, LBP provide a feature vector with 256 dimensions and LTP extract 2×256 features. In this step, extracted feature vectors are concatenated together. Next, second order statistics, which is called GLCM, would be estimated for the original image. GLCM can be estimated in 8 directions and achieve one matrix in every direction. Finally, to achieve discriminative features, 4 different popular statistical features are computed ever each one of the built GLCM matrixes as follows:

$$\text{Energy} = \sum_{i,j=0}^{N-1}(K_{i,j})^2 \qquad (4)$$

$$\text{Contrast} = \sum_{i,j=0}^{N-1} K_{ij}\,(i-j)^2 \qquad (5)$$

$$\text{Homogeneity} = \sum_{i,j=0}^{N-1} {K_{ij}}\Big/{1 + (i-j)^2} \qquad (6)$$

$$\text{Entropy} = \sum_{i,j=0}^{N-1} -\text{Ln}(K_{i,j})K_{i,j} \qquad (7)$$

$$\text{Variance} = \sum_{i=1}^{N}\sum_{j=1}^{M}(i-j)^2\,K_{ij} \qquad (8)$$

$K_{ij}$ is the matrix value in the position (i,j) in the output GLCM matrix. N is the number of gray levels in the output image. So, 8×5 feature is computed using GLCM. In this respect, totally 808 numerical features is extracted for each input image. To evaluate the texture classification, it is enough to use a numeric classifier. In the results part, it has been shown that by using these features, texture classification could be done by high accuracy.

## 4. Results

The main goal of this article is to propose an efficient approach for texture classification. In this respect a multi stage approach is proposed based on combination of LBP, LTP and GLCM features. In this section, to evaluate the performance of the proposed approach, Brodatz dataset is used as one of the state-of-the-art datasets in texture analysis[20]. In order to evaluate performance, different classifiers are performed such as K-nearest Neighbor (KNN), naïve Bayes and Random forest. Results are shown in the Table 1. As it can be seen, KNN with K=3 provides maximum accuracy between classifiers. In all experiments, 50 percent of the brodatz samples are used as train set and 50 percent another images as test set. Some image samples of Brodatz dataset is shown in the figure 5.

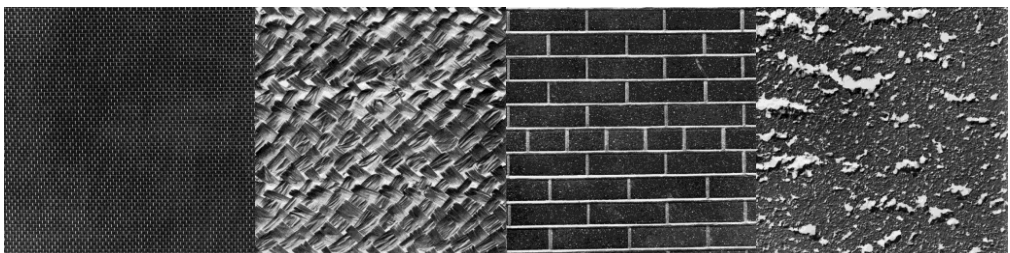

**Figure 5.** Some image samples of Brodatz dataset

**Table1**. Performance evaluation of proposed approach using different classifiers in terms of accuracy (%)

| Classification model / Approach | KNN (K=1) | KNN (K=3) | KNN (K=5) | Naive Bayes | Random forest |
|---|---|---|---|---|---|
| *Proposed approach* | 89.33 | 92.83 | 92.24 | 78.89 | 88.63 |

Since now, many different approachs have been proposed for texture classification in gray-level images. So, the performance of our proposed approach is compared with some of state-of-the-art approachs in this scope. Compared results are shown in the table 2. As can be seen, in all cases, the proposed approach provides higher performance in terms of classification accuracy.

**Table2**. Performance evaluation of proposed approach in comparison with state-of-the-art approaches in terms of accuracy (%)

| Approach | $LBP_{8,1}$ | $LTP_{8,1}$ | GLCM | SEGL | SIFT | Proposed approach |
|---|---|---|---|---|---|---|
| *Accuracy* | 89.23 | 90.69 | 84.43 | 91.54 | 92.31 | 92.83 |

# 5. Conclusion

In this paper joint of gray level co-occurrence matrix in different directions and local ternary patterns are used for texture image classification. 768 numerical feature are extracted using LTP in three different color channels and 40 features are performed based on multi-direction GLCM. As results mentioned, our proposed provides acceptable accuracy in comparing with other state-of-the-art methods in this scope. Both of GLCM and LTP are two statistical texture analysis operators. So, the computational complexity of these operators is lower than deep networks. LTP is performed on whole of the image and GLCM is employed in different directions. Hence, the sensitivity of the proposed approach is lower than popular approaches in this scope. The extracted features is not sensitive to type of input image, so it can be used in different applications for learning phase too. The proposed approach can be used in many different computer vision applications such as image retrieval and object classification as future works.

# REFERENCES

[1] Zhang, J., & Tan, T. (2002). Brief review of invariant texture analysis methods. *Pattern recognition*, *35*(3), 735-747.

[2] Kociołek, M., Strzelecki, M., & Obuchowicz, R. (2020). Does image normalization and intensity resolution impact texture classification?. *Computerized Medical Imaging and Graphics*, *81*, 101716.

[3] Nanni, L., Lumini, A., & Brahnam, S. (2010). Local binary patterns variants as texture descriptors for medical image analysis. *Artificial intelligence in medicine*, *49*(2), 117-125.

[4] Fekri-Ershad, S. (2021). Cell phenotype classification using multi threshold uniform local ternary patterns in fluorescence microscope images. *Multimedia Tools and Applications*, *80*(8), 12103-12116.

[5] Verma, M., & Raman, B. (2018). Local neighborhood difference pattern: A new feature descriptor for natural and texture image retrieval. *Multimedia Tools and Applications*, *77*(10), 11843-11866.

[6] Verma, M., & Raman, B. (2015). Center symmetric local binary co-occurrence pattern for texture, face and bio-medical image retrieval. *Journal of Visual Communication and Image Representation*, *32*, 224-236.

[7] Kayhan, N., & Fekri-Ershad, S. (2021). Content based image retrieval based on weighted fusion of texture and color features derived from modified local binary patterns and local neighborhood difference patterns. *Multimedia Tools and Applications*, *80*(21), 32763-32790.

[8] Cheng, G., & Han, J. (2016). A survey on object detection in optical remote sensing images. *ISPRS Journal of Photogrammetry and Remote sensing*, *117*, 11-28.

[9] Lee, J. S., Kuo, Y. M., Chung, P. C., & Chen, E. L. (2007). Naked image detection based on adaptive and extensible skin color model. *Pattern recognition*, *40*(8), 2261-2270.

[10] Alsaffar, M. F. (2021). Elevation of Some Biochemical and Immunological Parameters in Hemodialysis Patients Suffering from Hepatitis C Virus Infection in Babylon Province. *Indian Journal of Forensic Medicine & Toxicology*, *15*(3), 2355.

[11] Pourkaramdel, Z., Fekri-Ershad, S., & Nanni, L. (2022). Fabric defect detection based on completed local quartet patterns and majority decision algorithm. *Expert Systems with Applications*, *198*, 116827.

[12] Rohrmus, D. R. (2005). Invariant and adaptive geometrical texture features for defect detection and classification. *Pattern Recognition*, *38*(10), 1546-1559.

[13] Hasan, A. H., Al-Kremy, N. A. R., Alsaffar, M. F., Jawad, M. A., & Al-Terehi, M. N. (2022). DNA Repair Genes (APE1 and XRCC1) Polymorphisms–Cadmium interaction in Fuel Station Workers. *Journal of Pharmaceutical Negative Results*, *13*(2), 32-32.

[14] Reddy, B. R., Mani, M. R., Sujatha, B., & Kumar, V. V. (2010). Texture Classification Based on Random Threshold Vector. *International Journal of Multimedia and Ubiquitous Engineering*, *5*(1).

[15] Suresh, A., Raju, U. S. N., & Kumar, V. V. (2010). An Innovative Technique of Stone Texture Classification Based on Primitive Pattern Units. *International Journal of Signal & Image Processing*, *1*(1).

[16] Liao, S., Law, M. W., & Chung, A. C. (2009). Dominant local binary patterns for texture classification. *IEEE transactions on image processing*, *18*(5), 1107-1118.

[17] Zheng, C., Sun, D. W., & Zheng, L. (2006). Recent developments and applications of image features for food quality evaluation and inspection–a review. *Trends in Food Science & Technology*, *17*(12), 642-655.

[18] Alsaffar, M. F., Haleem, Z., & Hussian, M. N. (2020). Study some immunological parameters for Salmonella Typhi patients in Hilla city. *Drug Invention Today*, *14*(2).

[19] Ojala, T., Valkealahti, K., Oja, E., & Pietikäinen, M. (2001). Texture discrimination with multidimensional distributions of signed gray-level differences. *Pattern Recognition*, *34*(3), 727-739.

[20] Ojala, T., Pietikainen, M., & Maenpaa, T. (2002). Multiresolution gray-scale and rotation invariant texture classification with local binary patterns. *IEEE Transactions on pattern analysis and machine intelligence*, *24*(7), 971-987.

[21] Nanni, L., Brahnam, S., & Lumini, A. (2011). Local ternary patterns from three orthogonal planes for human action classification. *Expert Systems with Applications*, *38*(5), 5125-5128.

[22] Armi, L., & Fekri-Ershad, S. (2019). Texture image Classification based on improved local Quinary patterns. *Multimedia Tools and Applications*, *78*(14), 18995-19018.

[23] Verma, M., & Raman, B. (2018). Local neighborhood difference pattern: A new feature descriptor for natural and texture image retrieval. *Multimedia Tools and Applications*, *77*(10), 11843-11866.

[24] Liu, L., Lao, S., Fieguth, P. W., Guo, Y., Wang, X., & Pietikäinen, M. (2016). Median robust extended local binary pattern for texture classification. *IEEE Transactions on Image Processing*, *25*(3), 1368-1381.

[25] Fekri-Ershad, S., & Ramakrishnan, S. (2022). Cervical cancer diagnosis based on modified uniform local ternary patterns and feed forward multilayer network optimized by genetic algorithm. *Computers in Biology and Medicine*, *144*, 105392.

[26] Haralick, R. M., Shanmugam, K., & Dinstein, I. H. (1973). Textural features for image classification. *IEEE Transactions on systems, man, and cybernetics*, (6), 610-621.

[27] Mohanaiah, P., Sathyanarayana, P., & GuruKumar, L. (2013). Image texture feature extraction using GLCM approach. *International journal of scientific and research publications*, *3*(5), 1-5.

[28] Alsaffar, M. F. (2019). Studying of certain immunological parameters in the Province of Babylon for systemic lupus erythematosus. *Drug Invention Today*, *12*(11).

[29] Bakheet, S., & Al-Hamadi, A. (2021). Automatic detection of COVID-19 using pruned GLCM-Based texture features and LDCRF classification. *Computers in Biology and Medicine*, *137*, 104781.

[30] Rezaei, M., Saberi, M., & Ershad, S. F. (2011, December). Texture classification approach based on combination of random threshold vector technique and co-occurrence matrixes. In *Proceedings of 2011 International Conference on Computer Science and Network Technology* (Vol. 4, pp. 2303-2306). IEEE.

[31] Eltahir, M. M., Hussain, L., Malibari, A. A., K Nour, M., Obayya, M., Mohsen, H., ... & Ahmed Hamza, M. (2022). A Bayesian Dynamic Inference Approach Based on Extracted Gray Level Co-Occurrence (GLCM) Features for the Dynamical Analysis of Congestive Heart Failure. *Applied Sciences*, *12*(13), 6350.